\documentclass[letterpaper, 10 pt, journal, twoside]{IEEEtran}

\IEEEoverridecommandlockouts                              % This command is only needed if 
\usepackage[pdftex]{graphicx}
\usepackage{amsmath}
\usepackage{amssymb}
\usepackage{diagbox}
\usepackage{makecell}
\usepackage{mathtools} % For \prescript and other math tools
\usepackage{url}
\usepackage{hyperref}
\usepackage{bm}
\usepackage{array}
\let\labelindent\relax
\usepackage{enumitem}
\usepackage{color,soul}
\usepackage{leftindex}
\usepackage{todonotes}
\usepackage{color,soul}
\usepackage{xcolor}

\usepackage{tensor}   % For tensor notation with better spacing
\usepackage{stackengine}
\newcommand\barbelow[1]{\stackunder[1.2pt]{$#1$}{\rule{.8ex}{.075ex}}}
\newcommand\baron[1]{\stackon[1.2pt]{$#1$}{\rule{.8ex}{.075ex}}}

\makeatletter
\newcommand{\newparallel}{{\mathrel{\mathpalette\new@parallel\relax}}}
\newcommand{\new@parallel}[2]{%
  \begingroup
  \sbox\z@{$#1T$}% get the height of an uppercase letter
  \resizebox{!}{\ht\z@}{\raisebox{\depth}{$\m@th#1/\mkern-5mu/$}}%
  \endgroup
}
\makeatother

\renewcommand{\arraystretch}{1.5}  % Increase row height

\title{\LARGE \bf Flying Calligrapher: Contact-Aware Motion and Force Planning and Control for Aerial Manipulation}

\author{Xiaofeng Guo$^{*1}$, Guanqi He$^{*1}$, Jiahe Xu$^{1}$, Mohammadreza Mousaei$^{1}$, Junyi Geng$^{2}$,\\Sebastian Scherer$^{1}$, Guanya Shi$^{1}$%
\thanks{This work was supported by CMU Manufacturing Futures Institute (MFI).}%
\thanks{$^{*}$Equal Contribution.}%
\thanks{$^{1} $Xiaofeng Guo, Guanqi He, Jiahe Xu, Mohammadreza Mousaei, Sebastian Scherer, and Guanya Shi are with the School of Computer Science, Robotics Institute, Carnegie Mellon University, Pittsburgh PA 15213, USA. Email:
        {\tt\footnotesize \{xguo2, guanqihe, jiahex, mmousaei, basti, guanyas\}@andrew.cmu.edu}}%
\thanks{$^{2} $Junyi Geng is with the College of Engineering, Department of Aerospace Engineering, Pennsylvania State University, University Park, PA, 16802, USA. Email:
        {\tt\footnotesize jgeng@psu.edu}}%
}

\begin{document}

\maketitle
\thispagestyle{empty}
\pagestyle{empty}

%%%%%%%%%%%%%%%%%%%%%%%%%%%%%%%%%%%%%%%%%%%%%%%%%%%%%%%%%%%%%%%%%%%%%%%%%%%%%%%%
\begin{abstract}

Aerial manipulation has gained interest in completing high-altitude tasks that are challenging for human workers, such as contact inspection and defect detection, etc. Previous research has focused on maintaining static contact points or forces. This letter addresses a more general and dynamic task: simultaneously tracking time-varying contact force in the surface normal direction and motion trajectories on tangential surfaces. We propose a pipeline that includes a contact-aware trajectory planner to generate dynamically feasible trajectories, and a hybrid motion-force controller to track such trajectories. We demonstrate the approach in an aerial calligraphy task using a novel sponge pen design as the end-effector, whose stroke width is positively related to the contact force. Additionally, we develop a touchscreen interface for flexible user input. Experiments show our method can effectively draw diverse letters, achieving an IoU of 0.59 and an end-effector position (force) tracking RMSE of 2.9 cm (0.7 N). Website: \url{https://xiaofeng-guo.github.io/flying-calligrapher/}

\end{abstract}


%%%%%%%%%%%%%%%%%%%%%%%%%%%%%%%%%%%%%%%%%%%%%%%%%%%%%%%%%%%%%%%%%%%%%%%%%%%%%%%%

\section{INTRODUCTION}
\label{sec:introduction}

Uncrewed Aerial Vehicles (UAVs) have rapidly developed in recent years, with applications ranging from photography~\cite{bonatti2020autonomous, zhao2021super} to package delivery~\cite{geng2020cooperative, geng2021estimation}. High-altitude tasks, which are often challenging and time-consuming for human workers and sometimes require active interaction with the environment, have expanded the requirements for UAV operations beyond just passive roles. Aerial interaction, or aerial manipulation, where UAVs intentionally interact with the external environment, has gained increased attention. Such a shift has sparked greater interest in UAVs for tasks such as contact defect inspection, cleaning, or maintenance at high altitudes. The key challenge lies in the nonlinear floating-based disturbance~\cite{o2022neural,shi2019neural} as well as the coupling between the flying vehicle and the attached manipulator.

% why time-varying is import
There are many research efforts exploring the aerial manipulation in different scenarios, such as pushing a target~\cite{lee2020aerial}, turning a valve~\cite{brunner2022planning}, opening a door~\cite{lee2020aerialopendoor}, drilling~\cite{ding2021design}, screwing~\cite{schuster2022automated}, cleaning~\cite{alkaddour2023novel}, contact-based inspection~\cite{guo2023aerial, bodie2019omnidirectional}, etc. Uncrewed Aerial Manipulators (UAMs) in these scenarios often need to track a reference contact force or motion. However, most existing work focuses on specific tasks, such as maintaining a constant force or tracking discrete force targets at different locations. A general solution for tracking continuous, time-varying motion and force trajectories is lacking. On the other hand, there are many high-altitude applications that need this capability. For instance, in cleaning, the UAM's end-effector must slide on surfaces while adjusting to varying motion and contact force based on dirtiness. In carving or writing, the UAM must follow continuous, time-varying force and motion trajectories to accurately reflect stroke thickness. Relying solely on discrete force target tracking is too time-consuming and cumbersome.

\begin{figure}[!t]
    \centering
    \includegraphics[width = 0.99\linewidth]{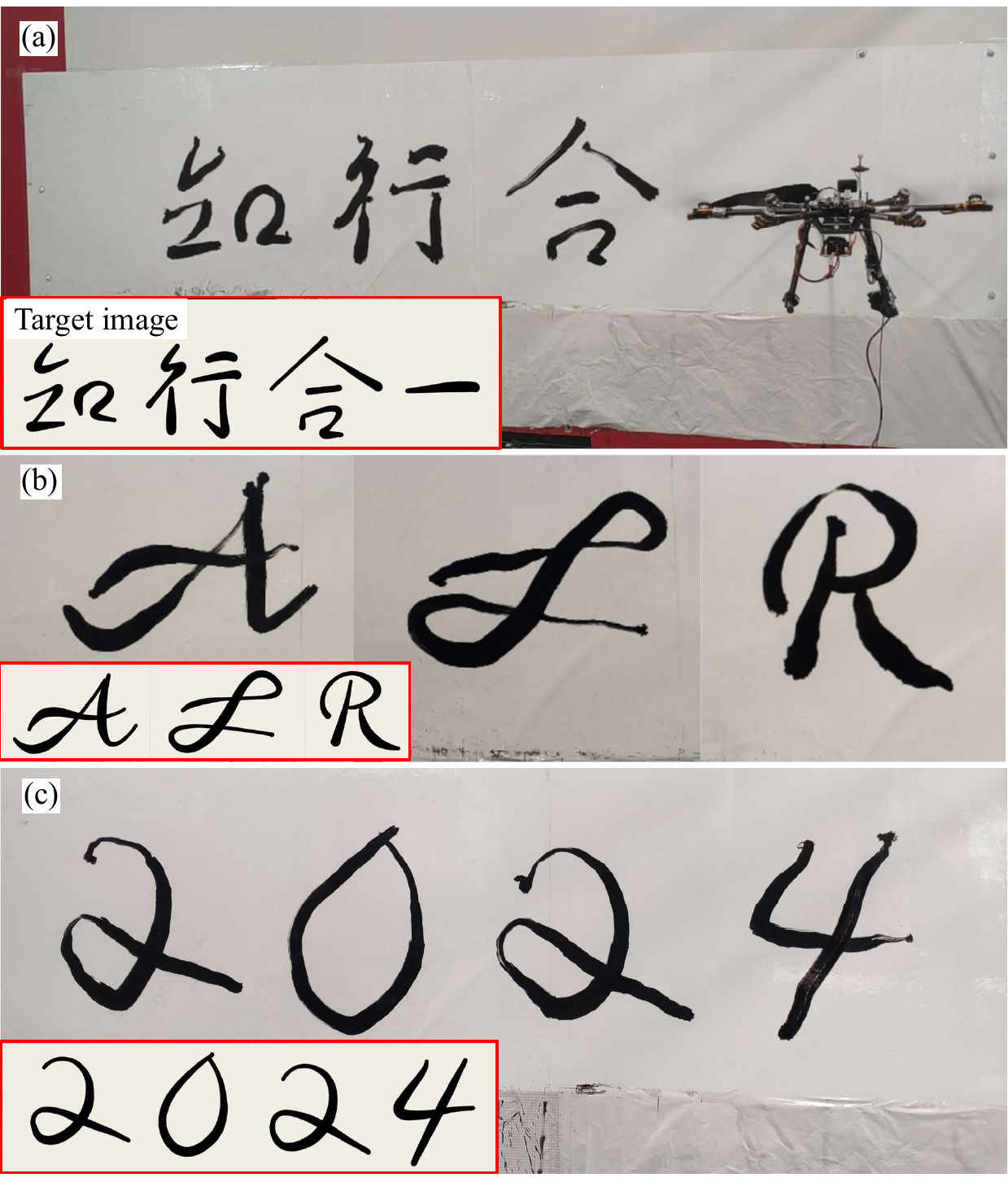}
    \caption{(a) Our aerial manipulator is writing a Chinese calligraphy artwork on the wall. (b) ``AIR'' and (c) ``2024'' written by our system.}
    \label{fig: concept}
\end{figure}

\begin{figure*}[]
    \centering
    \includegraphics[width = 0.99\linewidth]{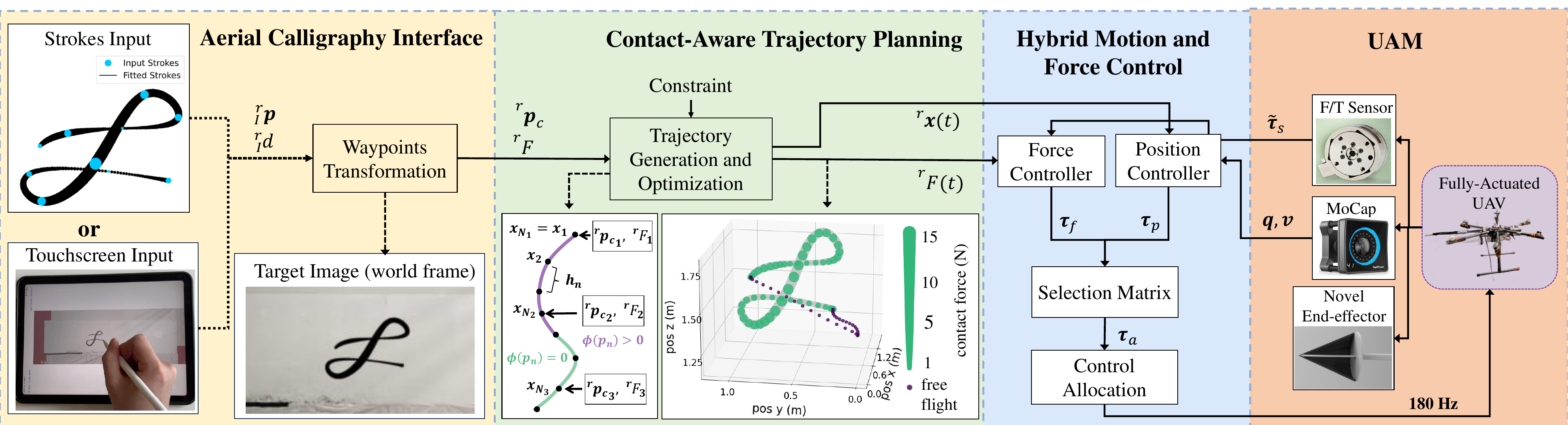}
    \caption{Pipeline of the proposed system. Users either manually define the target strokes or draw the target letter through a touchscreen interface. Then, the trajectory planning module generates a dynamic-feasible motion-force trajectory. A hybrid motion and force controller is designed for the UAM to track the reference trajectory. Finally, a novel end-effector pen is designed for the UAM to draw the target letters.}
    \label{fig: pipeline}
\end{figure*}

We aim to achieve general time-varying contact force and motion tracking to enable a wide range of aerial manipulation tasks. In particular, we target aerial calligraphy as a representative application to demonstrate the overall performance due to its high precision requirements on both motion and force tracking, and its superior visualization. Different from the simple aerial writing task, this task requires the UAM to draw letters with dynamically varying linewidth. It is challenging due to several reasons. The trajectory, including both the motion and contact force, must be smooth and dynamically feasible to ensure both feasibility and artistic quality. Unlike discrete force holding at different locations, the friction caused by the sliding with continuous time-varying contact forces during drawing must be properly handled by the controller. Additionally, the end-effector must be designed to reflect time-varying motion and contact force for proper visualization.

This paper develops a contact-aware hybrid motion-force planning and control strategy for aerial calligraphy using a fully-actuated hexarotor aerial manipulator. The system consists of three main components: a trajectory planning module, a hybrid control module, and a real-time user input interface via touchscreen. The touchscreen interface allows users to manually input or draw the target letter using an iPad. The trajectory planning module then generates a dynamically feasible motion and contact force trajectory, while the control module ensures the UAM accurately tracks these references to draw the letters. Additionally, a novel sponge pen end-effector is designed to visualize the drawn letters with varying linewidths based on contact force. In summary, the main contributions of this work are:

\begin{itemize}[leftmargin=*]
\item We propose a contact-aware trajectory planning algorithm to generate dynamically feasible contact force and motion trajectory in the contact plane.

\item We develop a contact-aware hybrid motion-force control algorithm to enable the UAM to track the continuous time-varying contact force and motion reference simultaneously, while compensating for friction force.

\item We develop a system and pipeline for a novel aerial calligraphy task. 
\end{itemize}

\section{RELATED WORKS}
\label{sec: related works}

In aerial manipulation tasks, UAMs often need to track reference contact force or motion trajectories during contact with external objects. To generate the reference trajectory, the existing works mainly focused on motion planning and treated the interaction force between the UAM and the environment as disturbances~\cite{fishman2021control, wang2024millimeter, lee2016planning}. As for contact force estimation, researchers have developed contact mechanics models, such as quasi-static or spring-damper models~\cite{brunner2022energy, zhang2022learning}. Some studies also mounted F/T sensors on UAMs to directly measure the contact force for feedback~\cite{peric2021direct, bodie2020active}. 

Aerial writing, which requires the UAM to control both motion and contact force during the task, is one common task the researcher focuses on. Tzoumanikas et al.~\cite{tzoumanikas2020aerial} developed a hybrid motion-force MPC as the trajectory generator and controller to draw letters with constant linewidth. They obtained the contact force using a linear spring model for vehicle dynamics. Lanegger et al.~\cite{lanegger2022aerial} designed a compliant Gough-Stewart mechanism for line drawing, improving stability and precision with multiple contact points. They equipped three omni-wheels with the end-effector and modeled the whole structure as a three-wheeled omnidirectional ground robot for precise control.

However, among all these works, most of them either track only a reference motion trajectory~\cite{peric2021direct, tzoumanikas2020aerial}, a set of discrete constant force targets at different locations~\cite{bodie2020active}, or time-varying force but without motion~\cite {byun2024image}. Our goal is to achieve general time-varying contact force and motion tracking to enable various aerial manipulation tasks, such as aerial calligraphy with strokes of varying linewidths. 
Unlike many studies that consider only normal contact forces, we utilize the full contact model in this work to predict the contact wrench (both normal force and friction force). This model is used in both trajectory planning to generate dynamically feasible trajectories and in hybrid motion-force control with contact wrench compensation.  
Additionally, our method treats execution time also as a decision variable to be optimized in trajectory optimization~\cite{wang2022geometrically}, improving UAM task efficiency compared to algorithms that predefine the execution time~\cite{fishman2021control, su2023sequential}.
\section{PROBLEM DEFINITION AND METHOD OVERVIEW}
\label{sec: problem def and method}

\subsection{Problem Definition}

This work aims to simultaneously track motion and contact force for UAM during aerial interaction. The problem is defined as follows: control the UAM's end-effector to navigate through a sequence of contact points on a surface with a known shape, while applying the desired normal contact force at each contact point.
Specifically, we focus on the aerial calligraphy task. Given a list of waypoints in the image space, each specifying the position and linewidth of a point, the goal is to control the UAM to accurately draw the letter.

\subsection{System and Pipeline Overview}
\label{sec: pipeline}

Our UAM platform is adopted from our previous works \cite{guo2023aerial}, which is a fully-actuated hexarotor equipped with a one-link manipulator. An  F/T sensor is mounted at the base of the manipulator. The flying calligrapher system pipeline is shown in Fig. \ref{fig: pipeline} and consists of four components. First, we develop an aerial calligraphy interface allowing users to specify the target image by either manually defining waypoints or drawing the letter through a touchscreen interface. This module then converts the user input to sparse waypoints in the task space, including the contact point position and magnitude of the normal contact force. Second, based on the sparse waypoints, we develop a contact-aware trajectory planner that generates a dynamically feasible motion-force trajectory for the UAM to track. Subsequently, a hybrid motion-force controller is designed so that the UAM can follow the reference trajectory. Finally, we design a novel end-effector for the UAV to draw the target letters.

\subsection{Coordinate Frames and Notations}

\begin{figure}[t]
    \centering
    \includegraphics[width = 0.95\linewidth]{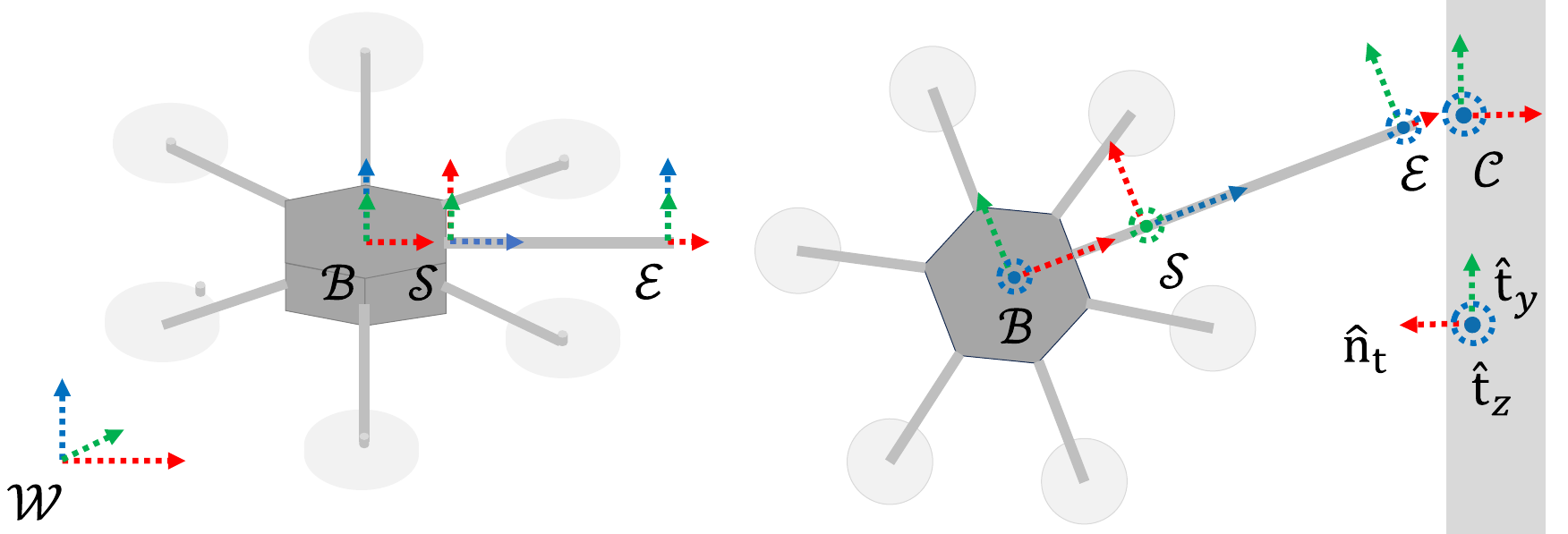}
    \caption{The coordinate frames used in this paper. Specifically, $\mathcal{W}$, $\mathcal{B}$, $\mathcal{E}$, $\mathcal{S}$, and $\mathcal{C}$ stand for the world, vehicle body, end-effector, arm, and contact frame, respectively.}
    \vspace{-0.2in}
    \label{fig: notations}
\end{figure}

We define several coordinate frames, as shown in Fig. \ref{fig: notations}. The world inertial frame  $\mathcal{W}$ is an east-north-up frame with the origin at the UAV's takeoff position. The body frame $\mathcal{B}$ originates at the vehicle’s center of mass, with axes pointing front, left, and up relative to the vehicle. The end-effector frame $\mathcal{E}$ originates at the end-effector, with axes pointing front, left, and up relative to the vehicle. The arm frame $\mathcal{S}$ originates at the root of the manipulator where the F/T sensor is mounted, with axes aligned to the sensor. The contact frame $\mathcal{C}$ originates at the contact point, with axes aligned to the surface normal (towards inside), left, and up directions of the surface.

Vectors are denoted as bold lowercase symbols. Left-hand subscripts, e.g., $_\mathcal{W}\bm{{v}}$, indicate the coordinate representation in frame $\mathcal{W}$. $\bm{R}^{b}_{a} \in \mathbb{R}^{3 \times 3}$, $\bm{T}_a^b \in \mathbb{R}^{3\times 1}$, $\bm{Ad}_{\bm{T}_a^b} \in \mathbb{R}^{6 \times 6}$ define the rotation, translation, and wrench (force and torque) transformation for vector from frame $b$ to frame $a$, respectively \cite{lynch2017modern}. We have: $_a\bm{v} = {\bm{R}_{a}^{b}} _b{\bm{v}}$. $\bm{0}_{a\times b}$ defines a $\bm{0}$ vector in $a \times b$ dimensions. $\bm{v}_{[i]}$ indicates the $i^{\mathrm{th}}$ element of $\bm{v}$; $^r(\cdot)$ denotes the reference quantity.

\subsection{Aerial Manipulator Dynamics}

We define the UAM end-effector position vector in the world frame as $_\mathcal{W}\bm{p}=[x, y, z]^\top$, the attitude with rotation matrix representation $\bm{R}_{\mathcal{W}}^{\mathcal{E}}$
and the angular velocity in the end-effector frame as $\bm{\omega}$. The pose and twist of the UAM end-effector are defined as $\bm{q} = [_\mathcal{W}\bm {p}, \bm{R}_{\mathcal{W}}^{\mathcal{E}}]$ and $\bm{v} = [_\mathcal{W}\dot{\bm{p}}^\top, \bm{\omega}^\top]^\top$.

The dynamic model of the hexarotor aerial manipulator at the end-effector can be derived following our previous work~\cite{he2023image}.

\begin{align}
\label{drone-model}
\bm{M}\dot{\bm{v}} + \bm{C}\bm{v} \!= \!{\bm{Ad}_{\bm{T}_{\mathcal{E}}^{\mathcal{B}}}} _\mathcal{B}\bm{\tau}_a + {\bm{Ad}_{\bm{T}_{\mathcal{E}}^{\mathcal{C}}}}_\mathcal{C}\bm{\tau}_c + {\bm{Ad}_{\bm{T}_{\mathcal{E}}^{\mathcal{B}}}} {\bm{R}_\mathcal{B}^\mathcal{W}}_\mathcal{W}\bm{g}
\end{align}with inertia matrix $\bm{M}\in\mathbb{R}^{6\times 6}$, centrifugal and Coriolis term $\bm{C}\in\mathbb{R}^{6\times6}$, gravity wrench $\bm{g}\in\mathbb{R}^{6}$, contact wrench $\bm{\tau}_c \in\mathbb{R}^{6}$, and control wrench $\bm{\tau}_a = [\bm f_a^\top, \bm m_a^\top]^\top \in\mathbb{R}^{6}$ from the UAM actuators, where $\bm f_a \in\mathbb{R}^{3}$ denotes the control force and $\bm m_a \in\mathbb{R}^{3}$ denotes the torque.
Specifically, we have

\begin{align}
\bm{M} = \begin{bmatrix}
    m\bm{R}_{\mathcal{E}}^{\mathcal{W}} & -m[_\mathcal{E}\bm{t}_{\mathcal{B}}^{\mathcal{E}}]_{\times} \\
    m[_\mathcal{E}\bm{t}_{\mathcal{B}}^{\mathcal{E}}]_{\times}\bm{R}_{\mathcal{E}}^{\mathcal{W}}& \bm{R}_{\mathcal{E}}^{\mathcal{B}}\bm{J}\bm{R}_{\mathcal{B}}^{\mathcal{E}} - m [_\mathcal{E}\bm{t}_{\mathcal{B}}^{\mathcal{E}}]_{\times}[_\mathcal{E}\bm{t}_{\mathcal{B}}^{\mathcal{E}}]_{\times}
\end{bmatrix}
\end{align}
\begin{align}
\bm{C} = \begin{bmatrix}
\bm{0}_{3\times3} & -m[\bm{\omega}]_{\times}[_\mathcal{E}\bm{t}_{\mathcal{B}}^{\mathcal{E}}]_{\times}  \\
\bm{0}_{3\times3} & [\bm{\omega}]_\times \bm{R}_{\mathcal{E}}^{\mathcal{B}}\bm{J}\bm{R}_{\mathcal{B}}^{\mathcal{E}} - m [_\mathcal{E}\bm{t}_{\mathcal{B}}^{\mathcal{E}}]_{\times}[\bm{\omega}]_{\times}[_\mathcal{E}\bm{t}_{\mathcal{B}}^{\mathcal{E}}]_{\times}
\end{bmatrix}
\end{align}

Here $m$ is the vehicle mass; $\bm{J}$ is the moment of inertia at the vehicle center of mass in the body frame; $_\mathcal{E}\bm{t}_{\mathcal{B}}^{\mathcal{E}}$ is the relative translation pointing from the tip of the end-effector to the vehicle body under the end-effector frame, and $[\bm{*}]_{\times}$ is the skew-symmetric matrix associated with vector $\bm{*}$.

The control wrench comes from the standard UAV rotor thrust and torque. As for the contact wrench, since the one-link manipulator is rigidly attached to the UAV contacting the surface through a point, only the reaction force is considered at the contact point without torque, i.e. 
\begin{align}
    _\mathcal{C}\bm{\tau}_c =[\bm{f}_c^\top, \bm{0}_{3\times1}^\top]^\top, \quad \bm{f}_c = \bm{f}_{\perp} + \bm{f}_{\newparallel} \label{eqn:contact_wrench}
\end{align}
where $\bm{f}_{\perp}$ and $\bm{f}_{\newparallel}$ represent the normal force and shear force respectively. Given the contact surface $\phi$ and the assumption that the end-effector makes a point-contact with the surface, we have the contact constraint $\phi(\bm{p})=0$. The contact axes can be represented by the surface normal vector $\hat{\bm{n}}_t$, and the in-plane basis vectors $\hat{\bm{t}}_y - \hat{\bm{t}}_z$ with 
\begin{equation}\label{contact-frame}
    \hat{\bm{n}}_t = \frac{\nabla\phi(\bm{p})}{\|\nabla \phi(\bm{p})\|}, 
    \hat{\bm{t}}_y = \frac{\bm{g}_{[1:3]}\times \hat{\bm{n}}_t}{\|\bm{g}_{[1:3]}\times \hat{\bm{n}}_t\|}, 
    \hat{\bm{t}}_z = \frac{\hat{\bm{t}}_y\times\hat{\bm{n}}_t}{\|\hat{\bm{t}}_y\times\hat{\bm{n}}_t\|}
\end{equation}

Therefore, the contact force can be expressed as

\begin{align}
    F &= \|\bm{f}_{\perp}\|, \quad  \bm{f}_{\perp} = F \bm{\hat{n}}_t\label{equation: shear_force}\\
    \bm{f}_\newparallel & = -\mu F\left(\text{sign}(\dot y) \hat{\bm{t}}_y + \text{sign}(\dot z) \hat{\bm{t}}_z\right) \nonumber\\
    \bm{\alpha}(\bm{v}) &= \left[ [\hat{\bm{n}}_t -\mu \text{sign}(\dot y) \hat{\bm{t}}_y - \mu \text{sign}(\dot z) \hat{\bm{t}}_z ] ^\top, \bm{0}_{3\times1}^\top \right]^\top\nonumber\\
    _\mathcal{C}\bm{\tau}_c &= F\bm{\alpha}(\bm{v})\nonumber
\end{align}
where $F$ is the magnitude of the normal force. Here, we apply the Coulomb model to estimate the friction force with the friction coefficient $\mu$. The negative sign at the beginning represents that the friction is in the opposite direction of the motion and $\text{sign}(\cdot)$ denotes the sign function.
\section{CONTACT-AWARE TRAJECTORY PLANNING}
\label{sec: planning}

This section introduces the proposed contact-aware trajectory planning algorithm. It takes $M$ sparse waypoints $({}^r{\bm {p}_c}_i, {}^rF_i), \text{for}\ i\in 1,\cdots, M$ and contact surface $\phi$ as input, and outputs the state and action trajectories $\bm x(t)$, ${\bm{\tau}_a}(t)$ as the reference for the hybrid motion-force controller, where $\bm x(t) = [{\bm{q}}^\top(t), {\bm v}^\top(t), {F}(t)]^\top$, $t \in (0, T)$. The waypoints can include both contact points where ${}^rF_i > 0$ and non-contact points where ${}^rF_i = 0$. Notice that our algorithm also optimizes the total time, therefore timestamps are not included in the input waypoints. Different from other works that consider contact reaction wrench as a disturbance, we treat $\bm{\tau}_c(t)$ as a predictable state in the trajectory planning, and we also output it to the controller. 

The trajectory planning algorithm is to find an N-step state and control sequences $\bm{x}_{1:N}$, ${\bm{\tau}_a}_{1:N}$, which can be formulated as an optimization problem:
% Objective
\begin{subequations}
\begin{align}
\label{opt-objective}\min_{\bm{x}_{1:N},\atop{{\bm{\tau}_a}_{1:N},\atop h_{1:N}}} &\sum_{n=1}^{N}  \left({\|\bm v_n\|^2}_{\bm W_v} + \|{{\bm \tau}_a}_n\|^2_{\bm W_\tau} + \|\delta {\bm \tau_a}_n\|^2_{\bm W_{d\tau}} + \gamma\right)h_n \\
s.t.\quad &\bm x_{n+1} = f_{dyn}(\bm x_n, {\bm \tau_a}_n,  h_n) \label{opt-dynamics} \\
    &0 < h_n \leq \bar{h} \label{opt-step} \\
    &\bm p_{N_i} = {^r\bm p_c}_{i},\quad F_{N_i} = {^rF}_{i}, \ \forall i \in 1, \cdots, M \label{opt-waypoint}\\
    & \phi({\bm {p}_{{n}}}) \begin{cases}
     =0, \text{if } {}^r{F}_{N_i}{}^r{F}_{N_{i+1}} > 0 \\
     >0, \text{if } {}^r{F}_{N_i}{}^r{F}_{N_{i+1}} = 0 \\
    \end{cases} \forall {{n}} \in [N_{i}, N_{i+1}] \label{opt-contact}\\
    & g(\bm x_n) = 0 \label{opt-task}\\
    &\barbelow{\bm{{\tau}}_a}\leq  {\bm \tau_a}_n \leq \bar{\bm{\tau}_a}, \quad
    \barbelow{\dot{\bm{{\tau}}}_a}\leq \delta {\bm \tau_a}_n \leq \baron{\dot{\bm{\tau}}_a} \notag\\
    &\barbelow{{\bm{v}}} \leq  {\bm v}_n \leq \bar{{\bm{{v}}}},\quad
\barbelow{\dot{{{F}}}} \leq \delta F \leq \bar{{\dot{F}}}\label{opt-force}
\end{align}
\end{subequations}

where $\|\bm x\|_{\bm W} := \bm x^\top \bm W \bm x$ denotes the $\bm W$-norm of $\bm x$. The cost function in \eqref{opt-objective} penalizes the twist $\bm v_n$, control wrench ${{\bm \tau}_a}_n$, changing rate of the wrench $\delta {\bm \tau_a}_n$ and time-interval between two steps $h_n$. $\bm W_v$, $\bm W_\tau$, $\bm W_{d\tau}$ are three diagonal weighting matrices and $\gamma$ is the scalar of time penalty. 

\eqref{opt-dynamics} represents the discrete-time dynamics of the system \eqref{drone-model}, derived using RK4 discretization. Since timestep $h_n$ is also an optimization variable, we impose an upper limit $\bar{h}$ on the time interval for each step~\eqref{opt-step} to ensure dynamic accuracy. The value of $\bar{h}$ is determined based on the trajectory length and the total number of steps $N$, and is fine-tuned during the experiments.

\eqref{opt-waypoint} ensures the UAM tracks waypoints by imposing the vehicle sequentially pass through the $i^\mathrm{th}$ waypoint at step $N_i$. Due to the upper limit for the time interval $h_n$, the optimization may become infeasible if there are insufficient steps between waypoints. Conversely, too many steps will cause a significant computational burden. We pre-allocate the step-index $N_i$ in proportion to the distance between adjacent waypoints to reduce the optimization time while ensuring feasibility. We set $N_1 = 1$ and $N = N_M$.

\eqref{opt-contact} ensures the end-effector contacts the target surface during the segment between two contact waypoints and ensures the end-effector does not make contact with the surface during other segments. 

\eqref{opt-task} provides the task-related constraint. Here, in the aerial calligraphy, to maintain the end-effector perpendicular to the wall for better drawing performance, the desired UAM orientation is constructed by \eqref{contact-frame}
\begin{align}
    g(\bm x) := {\bm{R}}_{\mathcal{W}}^{\mathcal{B}} \ominus {{}^r\bm{R}}_{\mathcal{W}}^{\mathcal{B}}, \quad {}^r{\bm{R}}_{\mathcal{W}}^{\mathcal{B}}(\bm p)& = [\hat{\bm{t}}_z, \hat{\bm{t}}_y, -\hat{\bm{n}}_t]
\end{align}

where $\ominus$ denotes the subtraction operation in the $SO(3)$ manifold. Additional safety and feasibility constraints for vehicle control wrench, velocity and contact force are added as \eqref{opt-force}, where $\barbelow{\bm{a}}$ and $\baron{\bm{a}}$ represents the lower limit and upper limit of the variable $\bm{a}$.

Since the time intervals obtained from problem \eqref{opt-objective} may be non-uniform, we apply linear interpolation to the planning results to ensure the control and state sequences are consistent with the controller frequency. Given $\bm x_{1:T}, {\bm{\tau}_a}_{1:T}$, $h_{1:T}$, for $t\in[T_j, T_{j+1}]$, and $T_j = \sum_{n=1}^{j} h_n$, 
\begin{align}
    \bm x(t) = \lambda \bm x_j + (1-\lambda) \bm x_{j+1}, \  {\bm \tau_a}(t) = \lambda \bm u_j + (1-\lambda) {{\bm \tau}_a}_{j+1} \nonumber
\end{align}    
with $\lambda = (t-T_j)/(T_{j+1}-T_j)$. We also calculate the model predicted contact wrench $\bm \tau_c (t)$ from \eqref{equation: shear_force} for later use. In summary, the planning algorithm outputs ${}^r\bm{x}(t)$, ${}^r\bm \tau_a(t)$, and ${}^r\bm\tau_c(t)$ for the hybrid motion-force controller.

\section{HYBRID MOTION-FORCE CONTROL}
\label{sec: control}

This section discusses the design of the hybrid motion-force controller. Based on our previous work~\cite{guo2023aerial}, we enhance the controller by incorporating contact wrench models, including the dynamics of contact force, friction force, and friction-induced torque. 

\subsection{Contact Wrench Estimation}
Unlike the free-flight UAV, the primary challenge of UAM is to compensate the contact wrench $\bm{\tau}_c$ during aerial interaction. We estimate it by fusing the sensor readings into the analytic model.

As described in \eqref{eqn:contact_wrench}, $_\mathcal{C}\bm{\tau}_{c}$ consists of both the normal and shear forces. Since we have an F/T sensor at the origin of frame $\mathcal{S}$, it allows us to estimate  $_\mathcal{C}\bm{\tau}_{c}$. 
Specifically, for the aerial calligraphy task, given the planned trajectory ${}^r{\bm{x}}$, we obtain ${}^r_\mathcal{C}\bm{\tau}_c = {}^r{F}{}^r{\bm{\alpha}}({}^r{\bm{v}})$. On the other hand, F/T sensor provides the measurement $_\mathcal{S}\Tilde{\bm{\tau}}_{s}$ for the wrench at the origin of frame $\mathcal{S}$ (denote as $_\mathcal{S}\bm{\tau}_{s}$). 
In addition, it is obvious to see that the relation of the $_\mathcal{S}\bm{\tau}_{s}$ and the contact wrench can be expressed as $_\mathcal{B}\bm{\tau}_c = {\bm{Ad}_{\bm{T}_{\mathcal{B}}^{\mathcal{S}}} }_\mathcal{S}\bm{\tau}_{s}$. 
Therefore, the contact normal force can be estimated by fusing the sensor measurement with the contact force model 
\begin{equation}
\centering
\hat{F} =  \left[ \bm{Ad}_{\bm{T}_{\mathcal{B}}^{\mathcal{C}}} {}^r\bm{\alpha}\right] ^{+} {\bm{Ad}_{\bm{T}_{\mathcal{B}}^{\mathcal{S}}}} _\mathcal{S}\Tilde{\bm{\tau}}_{s}
\label{equation: Fmea}
\end{equation}
where $[\cdot]^{+}$ denotes the pseudo-inverse of a matrix. Furthermore, under hover or zero tilt condition of the fully-actuated UAM and given the specific flat vertical wall, $\bm{R}_{\mathcal{B}}^{\mathcal{C}} \approx \bm{I}$. Thus, the estimate can be simplified as $\hat{F} = {[\bm{R}_{\mathcal{B}}^{\mathcal{S}}}{_\mathcal{S}\bm{\tau}_{s}}_{[1:3]}]_{[1]}$. The corresponding contact wrench estimates can be obtained as 
\begin{equation}
{\bm{_\mathcal{C}}\hat{\tau}_c} = \hat{F}\bm{\alpha}({}^r\bm{v})    
\end{equation}
Here, we use reference velocity ${}^r\bm{v}$ instead of measured velocity for robustness considerations.

\subsection{Hybrid Motion-Force Controller}

The fully-actuated UAM is able to control the translation and orientation independently. We develop a contact-aware hybrid motion-force controller based on our previous work \cite{guo2023aerial} for the UAM to track the planned trajectory ${}^r\bm{x}(t)$.

The control wrench $\bm{\tau}_a$ combines the output of the motion tracking controller $\bm{\tau}_p$ and wrench tracking controller $\bm{\tau}_f$ as follows:
\begin{equation}
    _\mathcal{B}\bm{\tau}_a = \bm{R}_\mathcal{B}^\mathcal{C}( ({\mathbf{I}_{6\times 6}}-\bm{\Lambda}){\bm{R}_{\mathcal{C}}^{\mathcal{B}}}_\mathcal{B}\bm{\tau}_{p} + \bm{\Lambda} _\mathcal{C}\bm{\tau}_f)
    \label{equation:tau}
\end{equation}
where $\bm{\Lambda} = \text{blockdiag}[\delta, 0, 0,0,0,0] \in\mathbb{R}^{6\times6}, \delta \in \{0, 1\}$. Intuitively, the matrix $\bm{\Lambda}$ selects direct force control commands and leaves the complementary subspace for the motion control. $\delta$ is the force motion control switch based on ${}^rF$. $\delta$ is set to $0$ when ${}^rF = 0$, corresponding to the pure motion control, and $\delta$ is set to $1$ when ${}^rF>0$, indicating the force control along the normal direction of the surface. 

For motion control, a PID controller is designed to directly control the end-effector motion.

\begin{align}
    _\mathcal{B} \bm \tau_p = &{_\mathcal{B}}^r\bm \tau_a + {\bm{Ad}_{\bm{T}_{\mathcal{B}}^{\mathcal{C}}}}({_\mathcal{C}}^r{\bm{\tau}}_c - _\mathcal{C}\hat{\bm\tau}_c) \notag\\
    &- \bm K_q \bm e_q - \bm K_v e_v - \bm K_i \int \bm e_q dt 
 \label{equation: motion_controller}
\end{align} 
where
\begin{align}
\bm e_q &= [\bm e_p^\top, \bm e_{R}^\top]^\top, \hspace{0.05in}
\bm e_v = [\dot{\bm e}_p^\top, \bm e_{\omega}^\top]^\top, \hspace{0.05in}
\bm{e}_{p} = \bm{R}_{\mathcal{B}}^\mathcal{W}(_\mathcal{W}\bm{p} - {_\mathcal{W}}^r{\bm{p}}) \notag\\
\bm{e}_{R} &= \frac{1}{2}({{{}^r{\bm{R}}_{\mathcal{W}}^{\mathcal{B}}}}^\top {\bm{R}_{\mathcal{W}}^{\mathcal{B}}}- {\bm{R}_{\mathcal{W}}^{\mathcal{B}}}{{{}^r{\bm{R}}_{\mathcal{W}}^{\mathcal{B}}}}^\top)^\vee, \hspace{0.05in}
\bm{e}_{\omega} = \bm{\omega} - {}^r{\bm{\omega}} 
\label{equation: motion_controller_details}
\end{align}

$_\mathcal{B}^r\bm{\tau}_a$ and $_\mathcal{C}^r\bm{\tau}_c$ come from the planning. Here $(\cdot)^\vee$ is to extract a vector from a skew-symmetric matrix and $\bm{K}_q, \bm{K}_v, \bm{K}_i$ are the positive definite tunable controller gain matrices. Here $_\mathcal{C}\hat{\bm{\tau}}_c$ is the key novelty of the proposed controller that we incorporate contact wrench models.

As for wrench tracking, we design an impedance controller to ensure that the end-effector maintains the desired wrench. 
Given the contact force reference $F^r$ and the estimated contact normal force \eqref{equation: Fmea}, the controller is designed as:
\begin{align}
    e_f & = \hat{F} - {}^rF, \quad
    e_{x}  = {\bm{e}_{q}}_{[1]} \notag\\
    F_f & = {}^rF - m_{0}m_{d}^{-1}(K_e e_{x} +  D_{e} {\dot{e}_{x}})  \\
    & \hspace{0.1in}  - K_{f,p}e_f -  K_{f,i}\int{e_f dt} \notag \\
    & = {}^rF - K_{e,p} e_{x} - K_{e,d} \dot{e}_{x} - K_{f,p}e_f -  K_{f,i}\int{e_f dt} \nonumber
    \label{eqn:impedance}
\end{align}

with $m_0$, $m_d$ the actual and desired vehicle mass, respectively, $K_{e,p}$ is the normalized stiffness, $K_{e,d}$ is the normalized damping. Then, the control wrench $_\mathcal{C}\bm{\tau}_f = [F_f, \bm{0}_{1\times5}]^\top$ and $_\mathcal{W}\bm{\tau}_p$ will be passed through \eqref{equation:tau} to get the total control wrench. Finally, $_\mathcal{B}\bm{\tau}_a$ is distributed to each rotor through the control allocation process.

Notice that $\hat{\bm{\tau}}_c$ in \eqref{equation: motion_controller}, and $\hat{F}$ in \eqref{equation: Fmea} take the contact force, including the normal force, friction force and the induced torque into account. Therefore, the designed controller is a contact-aware hybrid motion-force controller.
\section{END-EFFECTOR AND INTERFACE DESIGNS}

\label{sec:Aerial Calligraphy}

In this section, we introduce the end-effector and user interface design for the aerial calligraphy task.

\begin{figure}
    \centering
    \includegraphics[width = 0.99\linewidth]{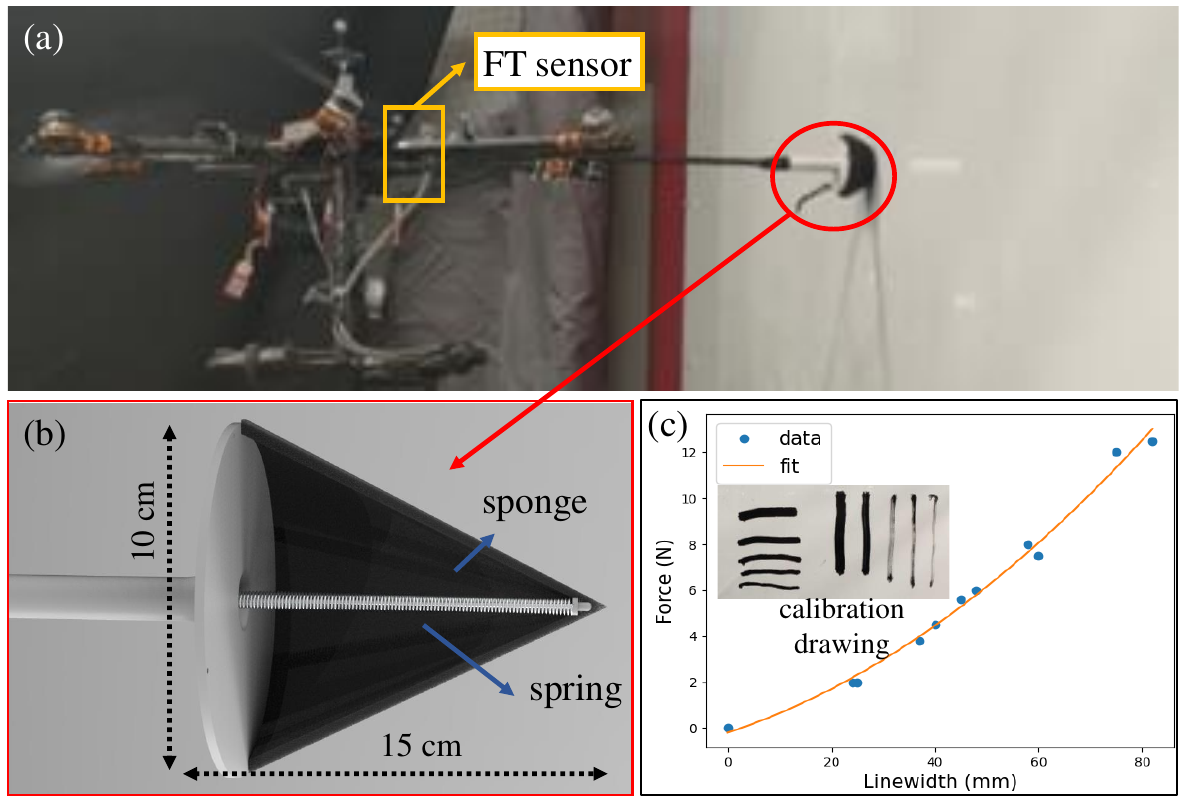}
    \caption{System overview. (a). Side-view of our UAM platform. (b). Design of the sponge pen as the end-effector. (c). Calibrated stroke linewidth and contact force relation.}
    \vspace{-0.2in}
    \label{fig:pen design}
\end{figure}

\subsection{End-Effector Design}
\label{sec:hardware}
Aerial calligraphy requires precise tracking of target linewidths, which can be achieved through contact force tracking. Most existing pens cannot produce varied linewidths, and specialized tools like the Chinese Calligraphy Brush need extra steps to align bristles for quality writing. In general, the pen for aerial calligraphy must satisfy these requirements: 1) Linewidth must correlate positively with contact force; 2) Changes in linewidth should be noticeable with varying force; 3) The correlation should be easily adjustable; 4) Writing should remain symmetrical during force application and release; 5) The pen should efficiently retain ink. We design a novel end-effector that meets these requirements.

Fig. \ref{fig:pen design} illustrates our pen design, consisting of two primary components: the sponge and the spring support. The cone-shaped sponge ensures isotropic properties and effectively absorbs and retains ink. A spring encircles a metal rod within the sponge, with adjustable stiffness and preload to fine-tune the force-linewidth curve. We dilute black ink with isopropyl alcohol (IPA) to enhance adhesion to the operating surface, which is a dry-erase whiteboard paper.

\subsection{Aerial Calligraphy Interface}
The interface allows users to input strokes from open-source datasets~\cite{gitrepo2018makehanzi} or through manual design. We also develop a touchscreen interface, implemented as an iPad application, which enables direct drawing of target strokes by capturing waypoints and the pressure applied during drawing. For each user input target strokes $s_i$, including waypoints $_\mathcal{I}\bm{p}_{i}$ (position) in image space $\mathcal{I}$ and $_\mathcal{I}d_i$ (linewidth), the interface processes it using Cubicspline fitting with constraints such as minimal curvature, and maximum and minimum linewidth to ensure feasibility. The target contact force $^rF_i$ is then calculated from the pre-calibrated force-linewidth relationship. The position $^r\bm{p}_{ci}$ is obtained by transforming $_\mathcal{I}\bm{p}$ to frame $\mathcal{W}$. For multi-stroke letters, we add the non-contact point at the endpoints of each stroke to enable stroke switching. Subsequently, the developed contact-aware trajectory planning algorithm generates the reference motion-force trajectory $\bm{x}(t), \bm{\tau}_a(t)$, which the UAM tracks to draw the target letter. In the real realization, we run the optimization for each free flight period and continuous contact period and then stitch them together.

\section{EXPERIMENT}
\label{sec: experiment}

In this section, we conduct experiments to demonstrate the effectiveness of our method. We perform aerial calligraphy on several letters from different languages and investigate the impact of various system modules and writing speeds on overall performance. 
More writing results are available on our project website.

\subsection{Experiment Setup}
\label{sec: experiment_setup}

We use a customized fully-actuated hexarotor UAV with onboard computation handled by an Nvidia Xavier AGX running ROS Noetic. System localization is achieved using an Optitrack Motion capture system. For aerial calligraphy, the newly designed pen (Sec. \ref{sec:hardware}) is mounted as the end-effector, and a dry-erase whiteboard paper is attached to a vertical wall for drawing. The friction coefficient is pre-measured as $\mu = 0.4$. The force-linewidth relationship is calibrated by drawing lines with varying contact forces and fitting a polynomial curve, as shown in Fig. \ref{fig:pen design}. Trajectory planning and optimization is performed using CasADI \cite{Andersson2018} with IPOPT \cite{wachter2006implementation} solver, initialized with a 0.1s time interval and linearly interpolated states between the waypoints, and hover thrusts. In real applications, we control the vehicle base for implementation convenience.

\renewcommand{\arraystretch}{1.2} % Adjust the number to your desired spacing

\begin{table}[]
\caption{Ablation Studies Results}
\centering
\label{table:comparison}
\begin{tabular}{>{\centering\arraybackslash}m{1.4cm} | >{\centering\arraybackslash}m{1.8cm} | >{\centering\arraybackslash}m{1.8cm} | >{\centering\arraybackslash}m{1.8cm}}
\hline
& Our Method & Control Baseline & Planning Baseline\\
    \hline
    EE Pos RMSE (cm) & \boldsymbol{$2.93$} $\pm$ \boldsymbol{$0.40$} & $5.47 \pm 0.74 $ & $4.40 \pm 0.14$ \\
    % \hline
    Base Pos RMSE (cm) & \boldsymbol{$2.48$} $\pm$ \boldsymbol{$0.39$} & $ 4.07 \pm 0.93$ & $3.68 \pm 0.65$\\
    Force RMSE (N) & \boldsymbol{$0.72$} $\pm$ \boldsymbol{$0.11$} & $1.89 \pm 0.07$ & $1.04 \pm 0.28$ \\
    \hline
    IoU & \boldsymbol{$0.59$} $\pm$ \boldsymbol{$0.09$} & $0.43 \pm 0.12$ & $0.49 \pm 0.06$ \\
    \hline    
\end{tabular}
\end{table}

\begin{figure}[h!]
    \centering
    \includegraphics[width = 0.99\linewidth]{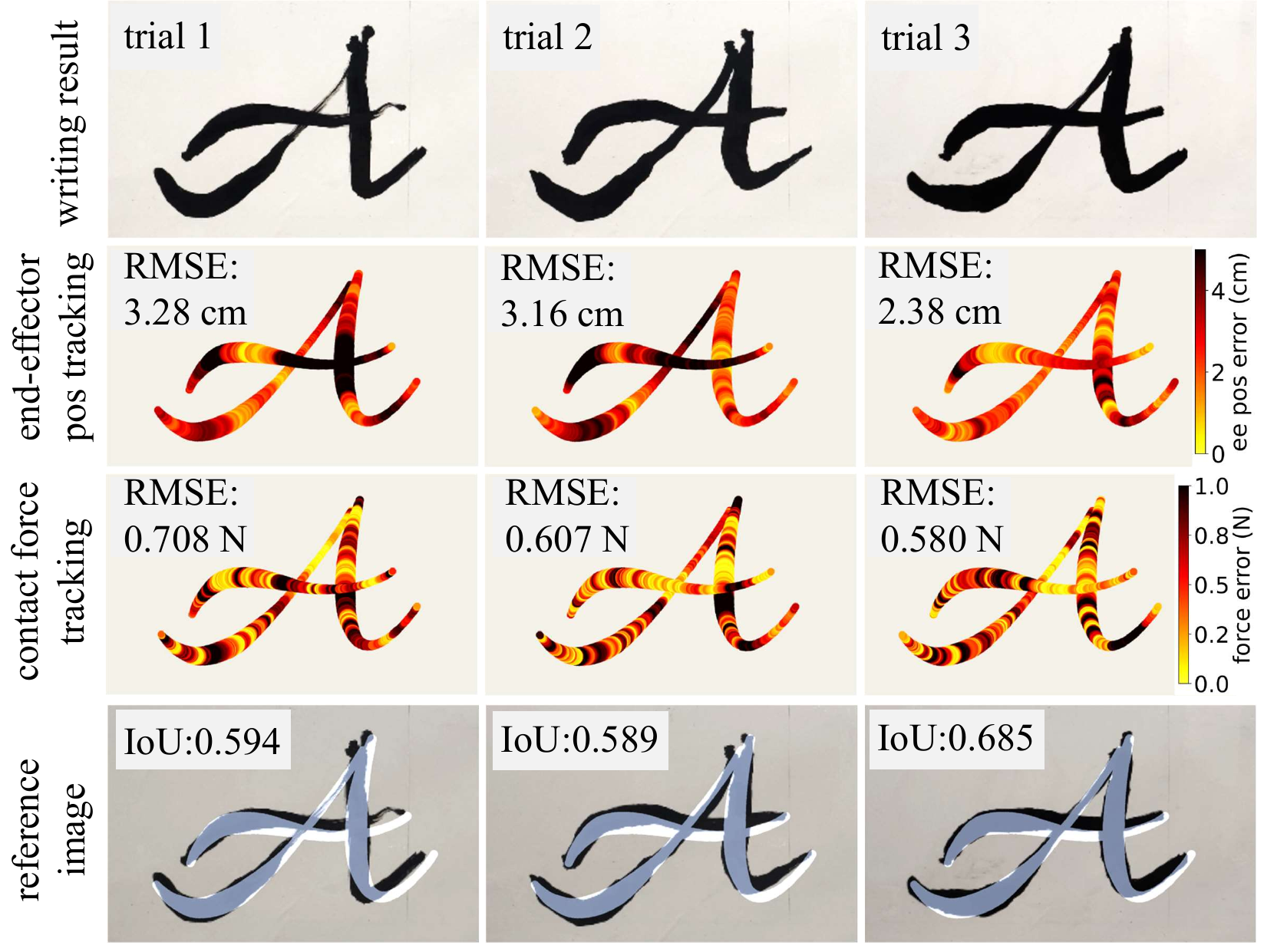}
    \caption{Results of three repeated trials for the letter `A'. Top row: Final written image. Middle two rows: Time history of end-effector position and contact force tracking performance. Bottom row: visual comparison between the written letter and ground truth image, evaluating the overall writing performance.}
    \label{fig: experiment_ourmethod}
\end{figure}

\begin{figure}[h!]
    \centering
    \includegraphics[width = 0.99\linewidth]{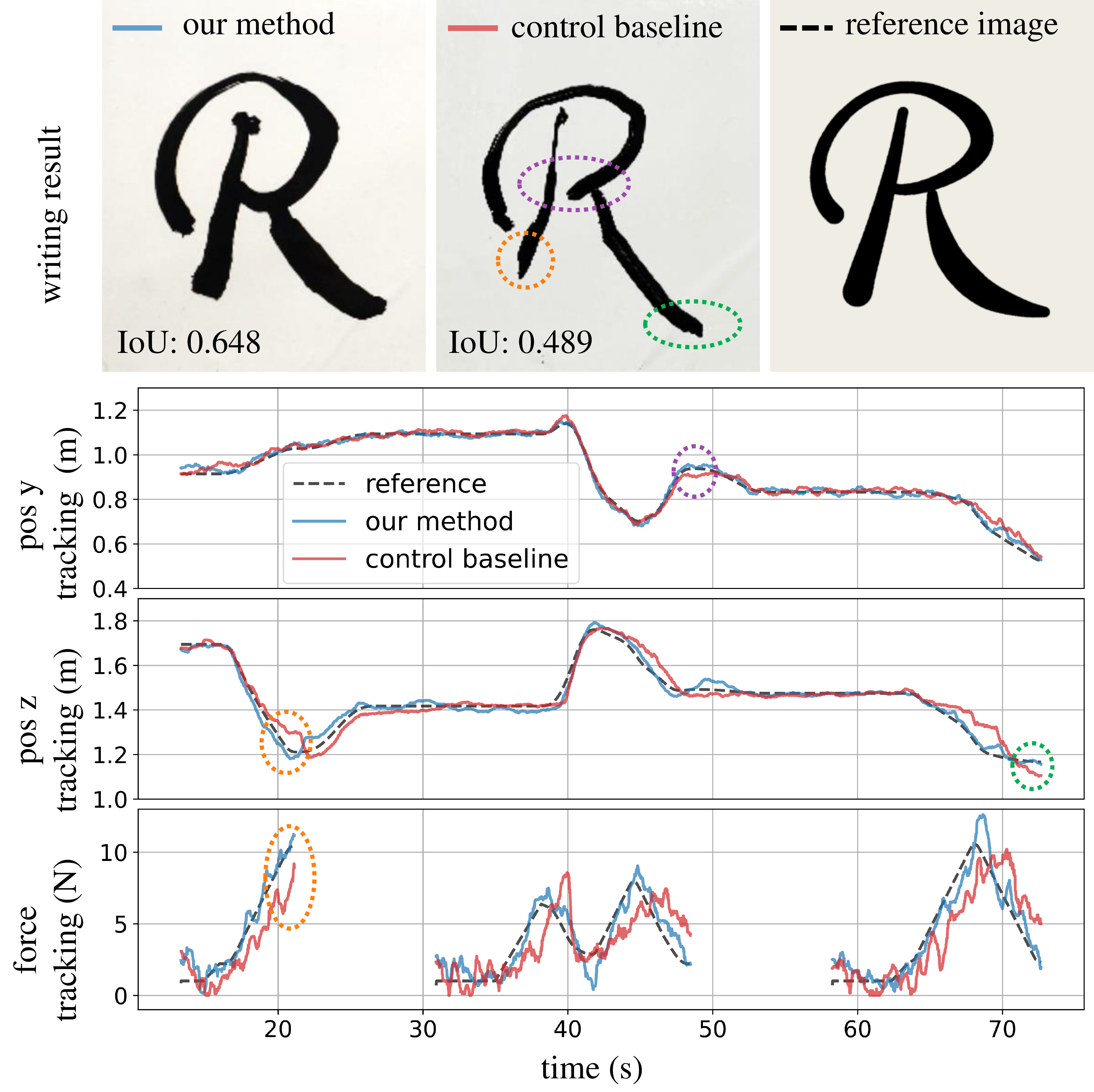}
    \caption{Comparison with the control baseline. Top: Writing results for letter `R'. Bottom: Time history of end-effector motion and contact force tracking performance.}
    \label{fig: experiment_control_comparison}
\end{figure}

\begin{figure}[h]
    \centering
    \includegraphics[width = 0.99\linewidth]{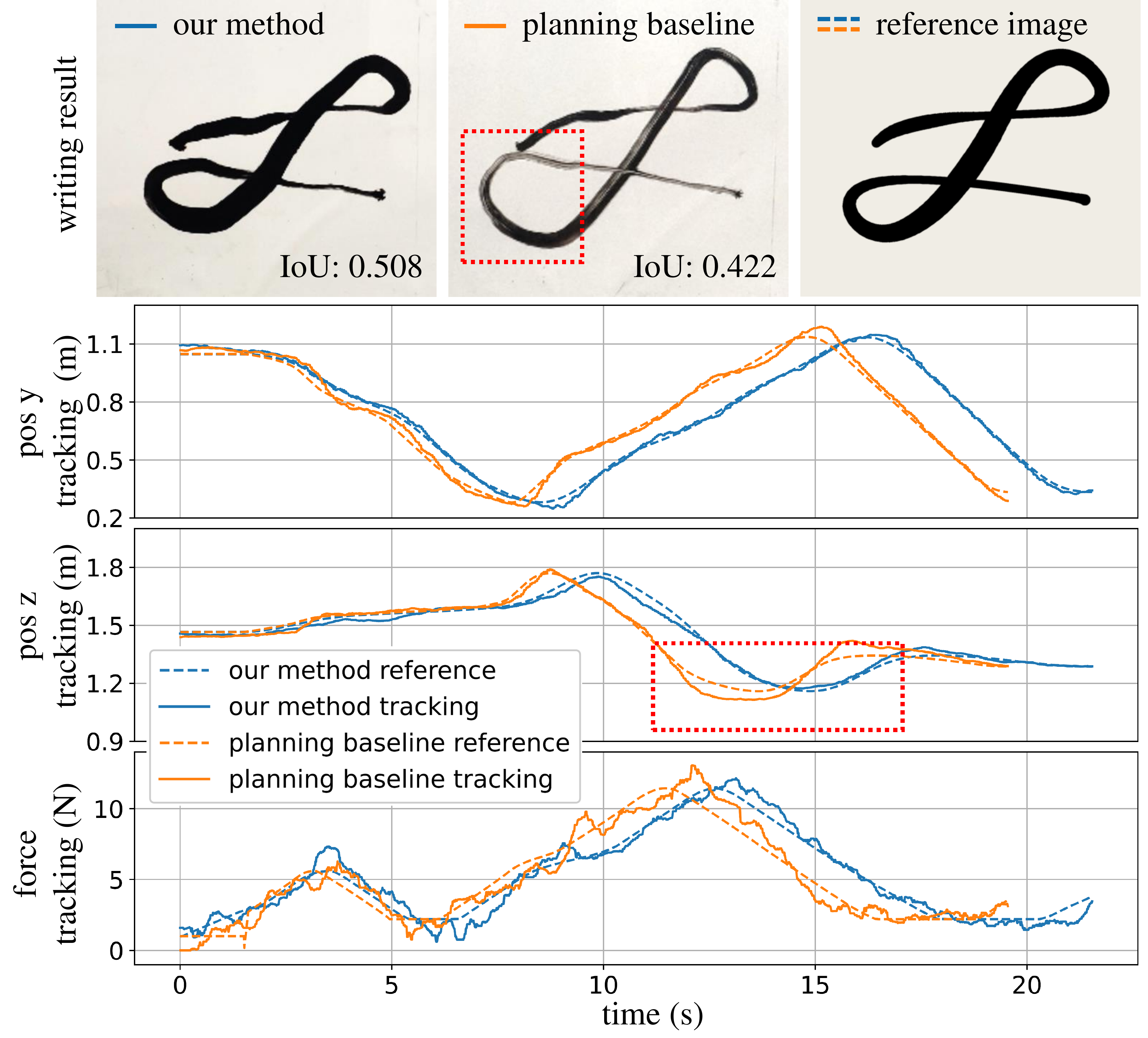}
    \caption{Comparison with the planning baseline. Top: Writing results for letter `I'. Bottom: Time history of end-effector motion and contact force tracking performance.}
    \label{fig: experiment_planning_comparison}
\end{figure}

\subsection{Aerial Calligraphy}

The aerial manipulator writes three English letters: `A', `I', and `R' in a custom typeface and font size. We manually define the sparse waypoints with point locations and local linewidths. The maximum reference velocity is set to $20~\mathrm{cm/s}$, the maximum acceleration to $20~\mathrm{cm/s^2}$; and the maximum contact force change rate to $2~\mathrm{N/s}$. We conducted three trials for each letter to test repeatability. 

Fig.~\ref{fig: experiment_ourmethod} shows the visual performance of the written letter. The UAM tracks motion and force trajectories well, with the written letters closely matching the target shapes and linewidths. Challenges arise when the UAM changes direction or at the start and end points of strokes due to the non-homogeneous property of the sponge pen during the establishment and detachment of contact. Smaller circular sections, such as the bottom-left curve of the letter `A', or the bottom-left part of the letter `I' (see Fig. \ref{fig: experiment_planning_comparison}), also present difficulties. This is presumably due to the rapid changing of friction mode (static or sliding friction) and friction force (both magnitude and direction) in these areas which requires a more accurate friction model. Despite these issues, the controller's performance is consistent across trials.

To quantitatively evaluate the performance, we calculate the Root Mean Squared Error (RMSE) by comparing the reference trajectory, shown in Tab. \ref{table:comparison}. The RMSE of the base position tracking is around $2.5$ cm and of end-effector (ee) position tracking is around $2.9$ cm. The contact force RMSE is around $0.7 \text{ N}$, which is better than the previous work that tracked a constant contact force~\cite{guo2023aerial}. This improvement is attributed to the developed contact-aware motion and force planning and control, which compensates for unmodeled behavior. 
We also use Intersection over Union (IoU) to evaluate the overall visual performance. We transform both the target and written letters into binary images, treating the letter as the target region at the pixel level. Our method achieves an average IoU of $0.59$, with an average intersection ratio of $0.79$ and an average union ratio of $1.35$.

We also test our system on a more complex scenario: Chinese idiom calligraphy. As shown in Fig.~\ref{fig: concept}, these characters feature more strokes, intricate shapes, and sharper turns. Our system still successfully wrote these characters with tracking and visual errors comparable to previous tests, highlighting the robustness of our approach across diverse writing tasks.

\subsection{Ablation Studies}
We conducted an ablation study to investigate the benefits of the proposed trajectory planning and hybrid motion-force control modules. We compared system performance under three scenarios: 1) our proposed approach; 2) without the friction and contact force model in the controller (Control Baseline); and 3) without contact-aware trajectory planning (Planning Baseline), using equal distance sampling of target waypoints for reference trajectory instead, which only passes $^r\bm{q}$ into the controller. The results are shown in Tab. \ref{table:comparison}.

\begin{figure}[!t]
    \centering
    \includegraphics[width = 0.99\linewidth]{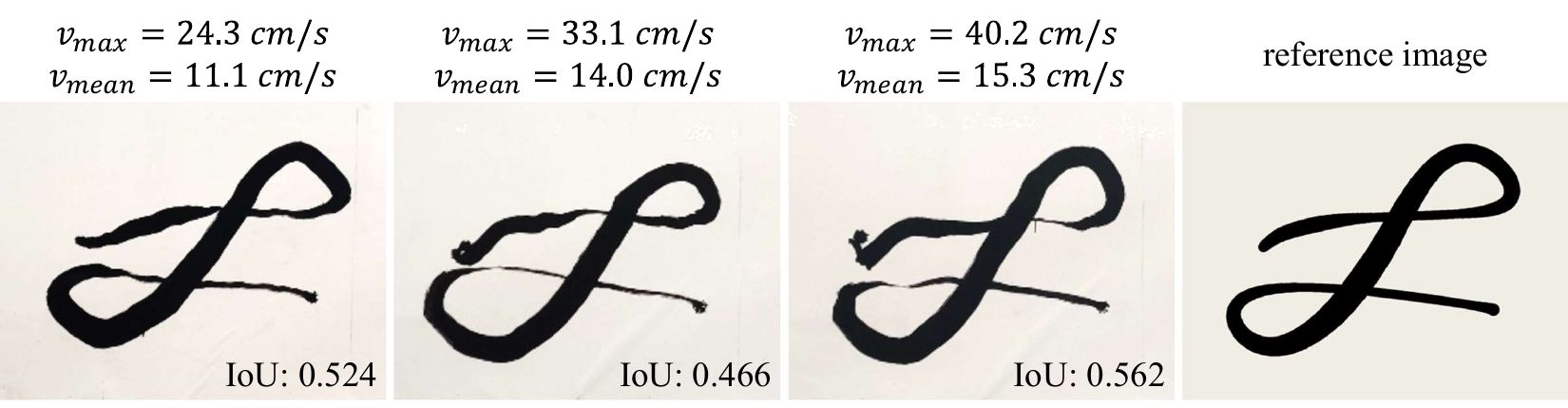}
    \caption{Visual results of aerial writing on the letter `I' under different velocity profiles.}
    \label{fig: experiment_speed_comparison}
\end{figure}

\begin{figure}[]
    \centering
    \includegraphics[width = 0.99\linewidth]{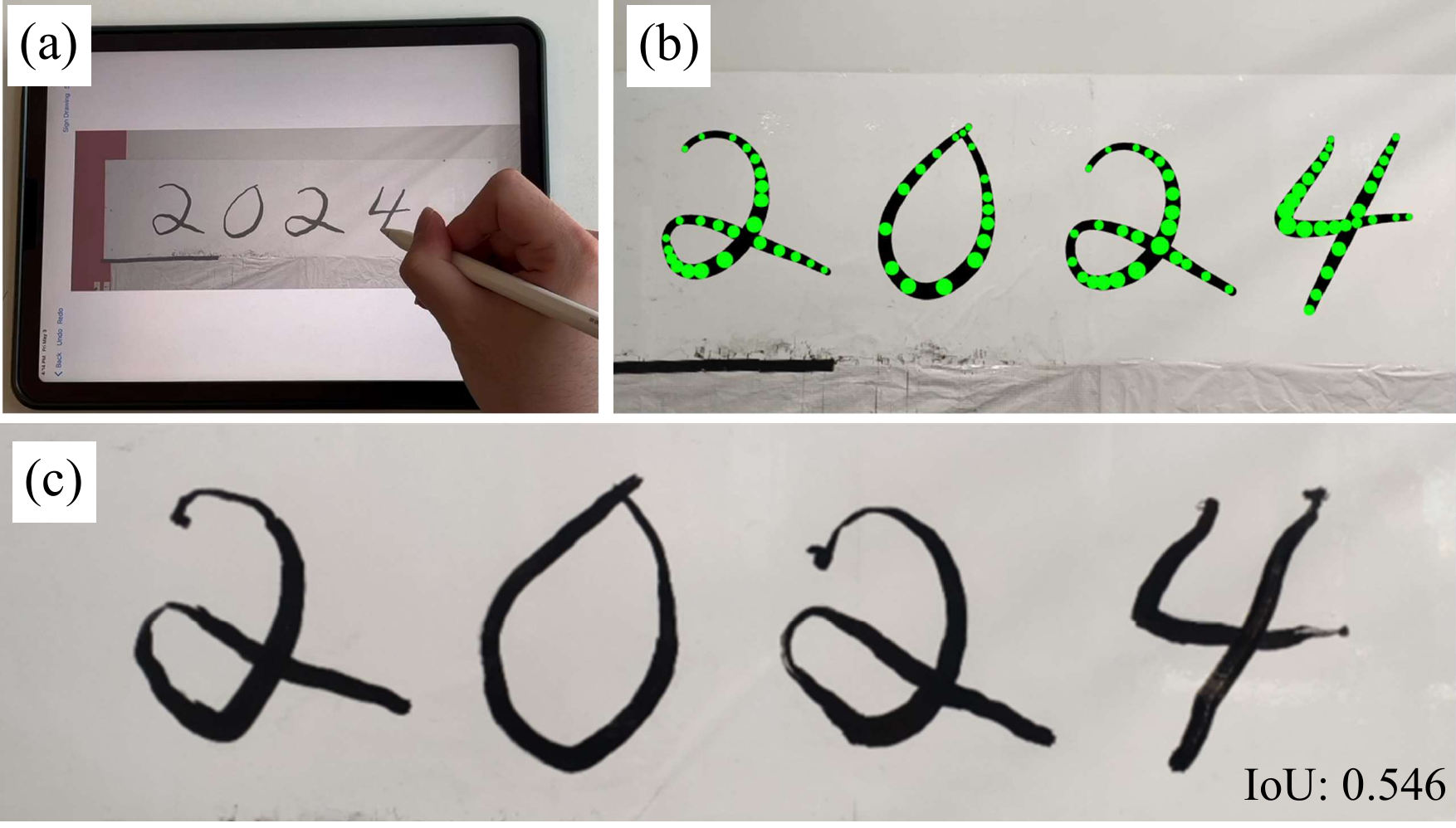}
    \caption{(a). The user inputs target strokes `2024' on the touchscreen. (b) The planned trajectory, where the green dots are the waypoints with equal timesteps. Regions with denser points indicate a slower planned velocity. (c). The actual letters written by the system.}
    \label{fig: experiment_IPAD}
\end{figure}

Fig. \ref{fig: experiment_control_comparison} shows the comparison with scenario 2). It is clear that without a proper contact force model, the contact force trajectory tracking is less accurate, leading to inconsistent stroke widths and occasional loss of contact. Without a proper friction force model, the written letter shapes are distorted, particularly at the ends of strokes and junctions, such as the vertical line and junction of the letter `R'.

Fig. \ref{fig: experiment_planning_comparison} shows the comparison with scenario 3). Without proper dynamic constraints, the linear interpolated reference trajectory makes sharp turns at corners, resulting in discontinuous velocity references that are unfeasible for the vehicle. This causes overshoots at corners, such as the bottom-left corner of the letter `I'. In addition, testing the Planning Baseline with a slower speed resulted in even worse performance due to frequent switching between static and sliding friction, introducing more complex friction dynamics.

\subsection{Speed Comparison}

To demonstrate the effects of the velocity on writing accuracy, we perform the experiments with three different velocity profiles whose maximum speed is 24.3 cm/s, 33.1 cm/s, and 40.2 cm/s. Fig. \ref{fig: experiment_speed_comparison} shows the visual results. The proposed method achieves a final writing IoU of 0.562 at a velocity of 40 cm/s. To the best of our knowledge, this is significantly faster than all related works~\cite{tzoumanikas2020aerial}. The main restriction on achieving higher velocities is the calibrated contact force change speed, limited by hardware and the contact model. Given that sliding at a slow speed introduces extra challenges, there is an optimal velocity range, highlighting the necessity of our trajectory planning algorithm.

\subsection{Aerial Calligraphy with Touchscreen Interface}
We demonstrate the convenience of our touchscreen interface for a complete aerial writing demo. A human user writes `2024' on the touchscreen. We then preprocess the input waypoints to make them feasible for our system. We regenerate the target image and render it for user verification. The UAM then performs contact-aware trajectory planning and hybrid motion-force control to draw the letters. Fig. \ref{fig: experiment_IPAD} shows the user's raw input, the regenerated target image, and the written letters by the UAM. The touchscreen provides a more intuitive way for users to input targets.

\section{CONCLUSION}
\label{sec: conclusion}
In this letter, we propose a general framework to simultaneously track time-varying contact force in the surface's normal direction and motion trajectories on tangential surfaces. Specifically, we propose an aerial calligraphy system. Our pipeline imports predefined target strokes or user drawings from a touchscreen, generates a dynamically feasible trajectory, and controls the UAM to write the target calligraphy. Experiments demonstrate that our method can draw diverse letters with effective motion and force tracking. The IoU of the final drawing is 0.59, and the end-effector position tracking RMSE is 2.9 cm. Future work includes integrating the touchscreen as an online teleoperation interface.

\addtolength{\textheight}{-12cm}   % This command serves to balance the column lengths

%%%%%%%%%%%%%%%%%%%%%%%%%%%%%%%%%%%%%%%%%%%%%%%%%%%%%%%%%%%%%%%%%%%%%%%%%%%%%%%%

%%%%%%%%%%%%%%%%%%%%%%%%%%%%%%%%%%%%%%%%%%%%%%%%%%%%%%%%%%%%%%%%%%%%%%%%%%%%%%%%


\section*{ACKNOWLEDGMENT}
This research is partially sponsored by the CMU Manufacturing Futures Institute (MFI). Any opinions, findings, conclusions, or recommendations expressed in this paper are those of the authors and do not necessarily reflect the views of the MFI.

\bibliographystyle{IEEEtran}
\bibliography{root}

\end{document}